\documentclass[conference]{IEEEtran}  

\usepackage[T1]{fontenc}
\usepackage{microtype}
\usepackage[dvipsnames]{xcolor}
\usepackage{graphicx}
\usepackage{amsmath,amssymb}
\usepackage[numbers,sort&compress]{natbib} 
\usepackage{algorithm}
\usepackage{algpseudocode}
\usepackage{threeparttable}
\usepackage{booktabs}
\usepackage{hyperref}
\hypersetup{colorlinks=true, linkcolor=MidnightBlue, citecolor=MidnightBlue, urlcolor=MidnightBlue}

\title{Using Variable Interaction Graphs to Improve Particle Swarm Optimization\\
\large (Extended version of the GECCO 2025 paper)}

\author{
\IEEEauthorblockN{Caz L. Czworkowski}
\IEEEauthorblockA{Johns Hopkins University\\
\texttt{cazczworkowski@gmail.com}}
\and
\IEEEauthorblockN{John W. Sheppard}
\IEEEauthorblockA{Montana State University\\
\texttt{john.sheppard@montana.edu}}
}

\begin{document}
\maketitle

\begin{abstract}
This paper presents Variable Interaction Graph Particle Swarm Optimization (VIGPSO), an adaptation to Particle Swarm Optimization (PSO) that dynamically learns and exploits variable interactions during the optimization process. PSO is widely used for real-valued optimization problems but faces challenges in high-dimensional search spaces. While Variable Interaction Graphs (VIGs) have proven effective for optimization algorithms operating with known problem structure, their application to black-box optimization remains limited. VIGPSO learns how variables influence each other by analyzing how particles move through the search space, and uses these learned relationships to guide future particle movements. VIGPSO was evaluated against standard PSO on eight benchmark functions (three separable, two partially separable, and three non-separable) across 10, 30, 50 and 1000 dimensions. VIGPSO achieved statistically significant improvements ($p<0.05$) over the standard PSO algorithm in 28 out of 32 test configurations, with particularly strong performance extending to the 1000-dimensional case. The algorithm showed increasing effectiveness with dimensionality, though at the cost of higher variance in some test cases. These results suggest that dynamic VIG learning can bridge the gap between black-box and gray-box optimization effectively in PSO, particularly for high-dimensional problems.
\end{abstract}

\textit{Extended version notice}—This is an extended version of the paper “Using Variable Interaction Graphs to Improve Particle Swarm Optimization,” GECCO 2025 (ACM), pp. 479–482. DOI: \href{https://dl.acm.org/doi/10.1145/3712255.3726589}{10.1145/3712255.3726589}.
\begin{IEEEkeywords}
Particle swarm optimization; variable interaction graphs; large-scale optimization
\end{IEEEkeywords}
\maketitle

\section{Introduction}
As previously presented in \citep{VIGPSO}, this paper extends those results substantially by introducing additional benchmark functions, higher dimensional analyses, and deeper theoretical insights. Particle Swarm Optimization (PSO) is a widely used optimization method for solving complex optimization problems. The concept was introduced as a method for solving nonlinear function optimization and has proven effective across a wide range of applications \citep{kennedy1995particle}. It is a population-based technique inspired by biological swarms, such as schools of fish or flocks of birds. The particles move through the solution search space and communicate to find optimal solutions. Each particle keeps track of the best solution it has encountered, while different PSO variants employ different communication topologies between particles, from fully connected networks to local neighborhoods \citep{comTops}. The balance between exploration and exploitation is managed by adjusting how much the particles move toward the global best solution versus exploring around their personal best solution or neighborhood best solutions.

Gray-box optimization involves enhancing algorithm performance by leveraging limited knowledge about the structure of the system being optimized. In black-box optimization, everything that happens between input and output depends solely on the input features. For example, native PSO is largely a black-box optimization method; however, gray-box optimization might be incorporated by including information about the problem space, such as the shape of the fitness landscape, or, as explored in this paper, the relationships between variables and their interactions.

When considering how to utilize information on variable interactions, the relationships between the variables can be captured in a Variable Interaction Graph (VIG), which stores known correlations between dimensions of the optimization task \citep{whitley2016gray}. The VIG provides partial knowledge about the interdependencies between variables. These graphs have demonstrated significant improvements in the performance of evolutionary algorithms \citep{10.1162/evco_a_00305}. Even so, in many cases, prior knowledge quantifying the interactions between different variables from which a VIG might be constructed is unavailable. In these situations, exploration during the training process can build these graphs. Previous work has examined incorporating problem landscape information into PSO \citep{van2004cooperative}, highlighting the challenges of high-dimensional problems. These approaches fall into the category of gray-box optimization, as they require prior knowledge that is not available in black-box problems.

This paper presents VIGPSO, a method for adapting VIGs to PSO for black-box optimization problems where nothing is known about the problem space. The primary advantage to the proposed approach is that the black-box nature of PSO incorporates information similar to a gray-box method by automatically using information on variable interactions discovered through the optimization process. The primary challenge lies in effectively creating an approximate VIG and leveraging it without any prior information about the underlying function. We hypothesize that incorporating a dynamically learned VIG into the PSO update equations will accomplish the following:
\begin{enumerate}
    \item Convergence rates will improve as compared to standard PSO variants on common benchmark functions, particularly in higher dimensions.
    \item The final solutions found will have statistically significantly lower objective function values ($p<0.05$), assuming minimization, across a range of problem types.
    \item More robust performance will be demonstrated across multiple runs, as evidenced by a lower standard deviation in the final solution quality, which reflects the algorithm's consistency in achieving similar results across different runs.
\end{enumerate}
The rationale behind these hypotheses is that the VIG serves as a form of memory for the model, storing its perceived variable interactions, which are unavailable in standard PSO. Prior research has shown that making PSO equations non-separable through dynamic parameter adaptation can improve both solution quality and convergence speed in high-dimensional problems \citep{HighDimSpace}. This principle supports the rationale for incorporating dynamically learned variable interactions into VIGPSO, as it enhances the algorithm's ability to navigate complex, high-dimensional search spaces.

\section{Background and Related Work}
\label{sec:related}

Incorporating information about the fitness landscape has been demonstrated previously to enhance stochastic search methods such as PSO.
Towers \textit{et al.} examined the process of adaptively learning aspects of the problem from a new perspective \cite{Towers2024.02.28.582468}. Specifically, their method focused on the fitness landscape itself and predicted its features using the population. Once this information was gathered across several iterations, it was then used to select the most high-performing parameters based on the predicted ruggedness of the fitness landscape. In their experiment, the authors utilized a neural network to predict properties of protein fitness landscapes. They found that fitness landscapes can be estimated and used to effectively guide directed evolution. In their approach, they estimated the fitness landscapes as 
\begin{equation*}
\mathbb{E}\left[F(p_k^i)|F(p_0^i)\right] - \bar{F} \approx e^{-\theta kK}(F(p_0^i) - \bar{F})\label{eq:1}
\end{equation*}
where $\bar{F}$ is the overall mean fitness of the landscape, $\theta$ is the mutation rate, $k$ is the number of generations, and $K$ is the number of interacting sites. 
$K$ is ultimately what is used for the measure of ruggedness.

Chicano \textit{et al.} focused on the use of a new gray-box crossover operator called Dynastic Potential Crossover (DPX) for pseudo-Boolean optimization \cite{10.1162/evco_a_00305}. The authors utilize VIGs to decompose the problem, but the information about variable interaction was is assumed to be available \textit{a priori}; it was not updated throughout the search process. While their work focused on crossover operators in evolutionary algorithms rather than PSO velocity updates, their methods for constructing and utilizing VIGs are relevant. Specifically, they described two approaches for identifying variable interactions: 1) constructing a co-occurrence graph by analyzing which variables appear together in subfunctions, and 2) using Fourier transforms to detect non-zero coefficients between variable pairs \citep{terras1999fourier}. Their findings suggest that when variable interactions can be captured and leveraged effectively, even if discovered dynamically, optimization performance can improve.

The VIG allowed the algorithm to explore the dynastic potential, or possible offspring of two parent solutions. By using the VIG, Chicano \textit{et al.} used a method inspired by the clique-tree algorithm in probabilistic graphical models \citep{10.1162/evco_a_00305,cliqueTree,pgm}. Specifically, they created a Recombination Graph by removing variables from the VIG that had the same values in both parent solutions. Then, they converted this new graph into a chordal graph by adding edges until every cycle of length four or more has a chord. That chordal graph was then turned into a clique tree in order to efficiently organize the exploration of the dynastic potential. Although the authors treated the VIG as given, they acknowledged that it can be constructed from problem structure when necessary \citep{10.1162/evco_a_00305}.

Two key findings directly influenced our experimental design and expectations. First, VIG utility diminishes when many variables interact because, as explained by \cite{10.1162/evco_a_00305}, problems with high nonlinear variable interaction (i.e., high epistasis) face exponentially growing computational costs for interaction exploration. In DPX, this manifested as larger cliques in the VIG, which subsequently required exponential time for optimal exploration. Second, the approach proved most effective with sparse edge density, with performance degrading as density increased, which they demonstrated through an NK-Landscape analysis.

Kennedy and Eberhart introduced Particle Swarm Optimization (PSO) as a method inspired by the social behavior of biological swarms, such as bird flocking and fish schooling \cite{kennedy1995particle}. Initially conceived as a social simulation, PSO evolved into an optimization algorithm by leveraging simple velocity adjustment rules to explore solution spaces effectively. The movement of particles in classic ``gbest'' PSO is guided by the following velocity update equation: 
\begin{equation*}
{v}_{ij} = \omega{v}_{ij} + c_1r_1({pbest}_{ij} - {x}_{ij}) + 
c_2r_2(gbest_j - {x}_{ij})\label{eq:2}
\end{equation*}
where $i$ is the particle index, $j$ is the dimension, ${v}_i$ is the velocity of particle $i$, $x_i$ is the current position of particle $i$, $pbest_i$ is the best position that particle $i$ has found so far, $gbest$ is the best position found by any particle, $r_1$ and $r_2$ are random numbers between 0 and 1, $c_1$ and $c_2$ are hyperparameters, and $\omega$ is an inertia weight. This equation allows particles to balance exploration and exploitation by considering both individual and group knowledge. This framework has been successfully applied to a wide range of optimization problems.

A recent study by Tin\'{o}s \textit{et al.} demonstrated how VIGs can be learned and utilized during the optimization process itself, rather than requiring them to be known \textit{a priori} \cite{VIGSnew1}. Their approach to pseudo-Boolean optimization, which they refer to as ``Iterated Local Search'' (ILS), showed that two variables $x_g$ and $x_h$ interact if:
\begin{equation*}
\ f(x,g,h) - f(x,h) \neq f(x,g) - f(x)\
\end{equation*}
where $f(x,g)$ represents the fitness after flipping bit $g$ in solution $x$, and $f(x,g,h)$ represents the fitness after flipping both bits $g$ and $h$. This provides a mechanism for detecting true variable interactions during the search process without requiring additional fitness evaluations beyond those already needed for optimization. They proved that this approach does not produce false positives and found that in their experiments it successfully identified over 97\% of true variable interactions. Although they may not have captured all variable interactions, they proved that the interactions they did identify were valid. The key benefit is that this learning happens naturally during optimization without requiring dedicated sampling or evaluation overhead.

This dynamic VIG learning method offered particular value for PSO adaptation to exploit variable interactions, providing a mathematically rigorous approach to dependency detection during swarm exploration.

Note that the idea of detecting and using information about variable interaction has become important in the larger context of cooperative co-evolutionary algorithms \citep{ccea94,van2004cooperative}. In particular, the development of the ``differential grouping'' strategy by Omidvar \textit{et al.}, applies a method for identifying groups of interacting variables using an approach similar to that of Tin\'{o}s \textit{et al.} \citep{DG-orig}. Specifically, the DG approach defines
\[
\Delta_{\delta,x_p}[f](\mathbf{x}) = f(\ldots,x_p+\delta,\ldots)-f(\ldots,x_p,\ldots)
\]
where $x_p$ is a variable, $\delta$ is an interval over which interaction is being tested, and $f(\cdot)$ is the function (e.g., the fitness landscape) being evaluated. Then $\forall a,b_1,b_2$ where $b_1\neq b_2$, then variables $x_p$ and $x_q$ are said to be non-separable (i.e., interacting) if
\[
\Delta_{\delta,x_p}[f](\mathbf{x})|_{x_p=a,x_q=b_1} \neq \Delta_{\delta,x_p}[f](\mathbf{x})|_{x_p=a,x_q=b_2}.
\]

Several DG variants have been proposed, including global differential grouping (GDG) \citep{GDG}, extended differential grouping (XDG) \citep{XDG}, dual differential grouping (DDG) \citep{DDG}, overlapping differential grouping (ODG) \citep{ODG}, and recursive differential grouping (RDG) \citep{RDG}. All of these methods, however, require the variable interactions to be determined before search begins. Perhaps most relevant to our work is the development of a \textit{dynamic} DG method that permits the groupings to be updated during the search process \citep{dyn-DG}.

\section{Optimization Strategy}
Unlike the ILS algorithm described by Tin\'{o}s \textit{et al.}, which used VIGs to make coordinated discrete changes to solutions \cite{VIGSnew1}, our approach leverages learned variable interactions to modify the continuous trajectories of particles through the search space. Our VIG is created dynamically in the black-box optimization setting. The initialization phase, as outlined in Steps 1--2 of Algorithm \ref{alg:vigpso}, begins by creating a zero-weighted adjacency matrix $G$ representing potential interactions between dimensions. This matrix stores learned correlations between variables as the optimization progresses. For each PSO iteration, the algorithm computes standard PSO position and velocity updates and uses these updates to calculate pairwise correlations based on the Pearson correlation between dimension updates. This coefficient, which quantifies the linear relationship between variables with values ranging from $-1$ to $+1$, has been shown to be effective in optimization tasks by grouping variables with similar evolutionary trends \citep{PearsonsCorr}. Its use reduces computational costs while enhancing adaptability, making it a practical tool for identifying variable dependencies in dynamic systems.

As shown in Steps 24--31 of Algorithm~\ref{alg:vigpso}, weak correlations below a pruning threshold $\tau_2$, are removed from the VIG. This ensures that only the stronger correlations are retained, reducing noise and emphasizing meaningful interactions. Specifically, the pairwise correlations between dimensions, computed from particle movements, are compared against the thresholds. Correlations above the pruning threshold are retained, while those falling below $\tau_2$ are removed to maintain the integrity of the graph.

\begin{algorithm}[t!]
\caption{VIGPSO}\label{alg:vigpso}
\begin{algorithmic}[1]
\State Initialize PSO parameters: particles, velocities, $pbest$, $gbest$
\State Initialize empty Variable Interaction Graph (VIG) matrix $G \gets 0$
\For{$t = 1$ to $max\_iterations$}
    \State Compute inertial weight $\omega_{curr} \gets \omega(1-0.6 prog$)
    \State Store current positions as $X_{old}$
    \For{each particle $p$}
        \For{each dimension $d$}
            \State Compute standard PSO velocity $v_s$
            \State Retrieve connected dimensions $\mathcal{N}$ from $G$
            \If{$\mathcal{N}$ is not empty}
                \State Get weights: $w_n \gets G_{d,n}$ for $n \in \mathcal{N}$
                \State Normalize weights: $w_n \gets w_n/\sum_{n \in \mathcal{N}} w_n$
                \State $v_{\text{vig}} \gets \sum_{n \in \mathcal{N}} w_n  v_n$
                \State $\alpha \gets 0.3 (1 - e^{-2t/t_{max}})$
                \State $v' \gets (1 - \alpha) v_s + \alpha v_{\text{vig}}$
            \Else
                \State $v' \gets v_s$
            \EndIf
            \State Clip velocity $v'$ to bounds
        \EndFor
        \State Update position $x \gets x + v'$
        \State Update $pbest$ and $gbest$ if improved
    \EndFor
    
    \If{$t \bmod update\_interval = 0$}
        \State Compute particle movement $\Delta X \gets X - X_{old}$
        \For{each pair of dimensions $i, j$}
            \State Compute correlation $\rho \gets corr(\Delta X_i, \Delta X_j)$
            \If{$|\rho| > \tau_1$}
                \State $G_{i,j} \gets |\rho|$
            \ElsIf{$|\rho| < \tau_2$}
                \State $G_{i,j} \gets 0$
            \EndIf
        \EndFor
    \EndIf
\EndFor
\State \Return $gbest$
\end{algorithmic}
\end{algorithm}

After updating the VIG, its information is incorporated into the PSO update equations by grouping correlated dimensions. This involves adjusting inertia or acceleration based on the VIG's data. As detailed in Steps 8--17 of Algorithm \ref{alg:vigpso}, the VIG information is incorporated into the PSO velocity update equation through a novel adaptive weighting scheme. For each dimension $d$, the standard PSO velocity update is combined with influences from connected dimensions in the VIG, as follows:
\begin{eqnarray*}
    v'_d &=& (1-\alpha)v_s + {\alpha}v_{\text{vig}}\\
    v_{\text{vig}} &=& \frac{\sum_{i \in \mathcal{N}_d} w_i v_i}{\sum{i \in \mathcal{N}_d} w_i}
\end{eqnarray*}
where $v'_d$ is the new velocity for dimension $d$, $v_s$ is the standard PSO velocity update result, $w_i$ is the edge weight from the VIG between $d$ and each member of the neighborhood $\mathcal{N}_d$, defined as those variables adjacent to $d$ in the VIG. The $v_{\text{vig}}$ is the VIG-based velocity component, and $\alpha$ is an adaptive weight that increases with iteration count to emphasize the influence of the VIG over time. 
Alternative communication topologies based on $lbest$, such as ring or star, can be used in place of the fully connected network implied by $gbest$.
These topologies may impact convergence behavior, particularly for high-dimensional or rugged fitness landscapes, and could interact with the VIG's adaptive weighting scheme. Note that the velocity magnitudes are clipped to prevent velocity explosions \citep{VelocityClamp}.

The adaptive weight $\alpha$, which determines the relative contribution of $v_{\text{vig}}$, evolves as optimization proceeds according to:
\begin{equation*}
    \alpha = 0.3(1 - e^{-2prog})
\end{equation*}
where $prog$ represents the relative progress ($t/max\_iterations$). This function was chosen to provide a smooth transition from exploration to exploitation, starting near 0 and asymptotically approaching 0.3 as $prog$ increases. The value of 0.3 was chosen empirically based on preliminary experiments, as it provided a balance between exploration and exploitation. The VIGPSO algorithm enhances standard PSO by incorporating learned variable interactions through this adaptive VIG mechanism. The use of $\alpha$ ensures that the influence of the VIG increases over time, improving the algorithm’s adaptability to the problem landscape. 


\section{Experimental Design}
We tested the effectiveness of our approach using benchmark functions specifically chosen to represent different types of variable interactions and separability characteristics. Three categories of functions were selected to help understand how the VIG adaptation performed under different optimization scenarios. For consistency, all functions were evaluated within the bounds $[-5,5]$ for all dimensions.
For fully separable functions where variables could be optimized independently, we used the Sphere function:
\[
f(x) = \sum_{i=1}^{n} x_i^2,
\]
the Sum Squares function:
\[
f(x) = \sum_{i=1}^{n} ix_i^2,
\]
and Schwefel 2.22:
\[
f(x) = \sum_{i=1}^{n} |x_i| + \prod_{i=1}^{n} |x_i|
\]
For partially separable functions, where groups variables interacted while the groups remained independent of each other, we used the Dixon-Price function:
\[
f(x) = (x_1-1)^2 + \sum_{i=2}^n i(2x_i^2 - x_{i-1})^2,
\]
and the Rastrigin function:
\[
f(x) = 10n + \sum_{i=1}^n [x_i^2 - 10\cos(2\pi x_i)].
\]
For fully non-separable functions, where all variables interacted in complex ways, we used the Rosenbrock function:
\[
f(x) = \sum_{i=1}^{n-1} [100(x_{i+1} - x_i^2)^2 + (1-x_i)^2],
\]
the Griewank function:
\[
f(x) = 1 + \sum_{i=1}^n \frac{x_i^2}{4000} - \prod_{i=1}^n \cos(\frac{x_i}{\sqrt{i}}),
\]
and the Alpine function:
\[
f(x) = \sum_{i=1}^n |x_i\sin(x_i) + 0.1x_i|.
\]

Performance was evaluated using dimensions of 10, 30, 50, and 1000 to test scalability while keeping computational requirements reasonable. Each algorithm configuration used 50 particles and ran for 300 iterations. For statistical validity, 100 independent runs were performed for each configuration. The key ways performance was evaluated were the solution quality (final objective value), 
convergence curves, robustness across the different function types noted above, and statistical significance testing between standard PSO and VIGPSO results using the Mann-Whitney U test (two-sided) with $\alpha$ = 0.05 significance level.

A comprehensive parameter sensitivity analysis was conducted separately for each of the 32 test configurations (8 functions for 4 dimensionalities). For each configuration, a grid search was performed over both PSO and VIGPSO parameters, including inertial weight {0.4, 0.5, 0.6, 0.8}, cognitive and social learning factors {1.0, 1.5, 2.0, 2.5}, correlation thresholds {0.3, 0.5, 0.7}, pruning thresholds {0.3, 0.5, 0.7}, and update intervals {5, 10, 15}. Each parameter combination was evaluated over 100 iterations to identify the best performing configuration for that specific test case. This function-specific tuning approach ensured both algorithms were operating with their optimal parameters among the grid points for each particular optimization scenario, providing a fair and rigorous comparison between the standard PSO and VIGPSO approaches.

The tuning results revealed that while optimal parameters varied across functions and dimensions, VIGPSO generally performed better with lower inertial weights ($\omega$=0.4) and higher social learning factors compared to cognitive factors, while standard PSO favored moderate inertial weights ($\omega$=0.6) with higher cognitive learning factors.

The time complexity of VIGPSO is $O(TSd^2)$ where $T$ is the number of iterations, $S$ is the number of particles, and $d$ is the number of dimensions. This is higher than standard PSO's complexity of $O(TSd)$  due to two main factors. First, the per-particle VIG influence computation requires $O(d^2)$ operations for each particle at every iteration. Here, each dimension potentially sums the velocity contributions from up to $d-1$ other dimensions. Second, updating the VIG itself has a complexity of $O(Sd^2)$, as the correlation calculation involves $O(S)$ operations for each pair of dimensions. While this additional computational overhead is significant, especially in high-dimensional spaces, the improved optimization performance often justifies the cost. The VIG update interval parameter allows some control over this trade-off, as updates only occur every $k$ iterations, though this does not affect the asymptotic complexity. 

\section{Results and Discussion}
As mentioned above, the benchmark experiments were conducted across eight test functions, categorized by their separability characteristics. The results of the Mann-Whitney U test are presented in Table \ref{tab:statistical_results}.
The column labeled ``Lower Obj.'' identifies the algorithm that returned the lower objective value (on average). A dash (``--'') indicates no statistically significant difference in performance.

\begin{table}[t!]
\centering
\caption{Statistical Comparison of VIGPSO vs Standard PSO}
\begin{tabular}{llccc}
\hline
\textbf{Func. Type} & \textbf{Function} & \textbf{Dim} & \textbf{p-value} & \textbf{Lower Obj.} \\
\hline
Separable & Sphere & 10 & 0.00000 & VIGPSO \\
& & 30 & 0.00000 & VIGPSO \\
& & 50 & 0.00000 & VIGPSO \\
& & 1000 & 0.00000 & VIGPSO \\
\cline{2-5}
& Sum Squares & 10 & 0.00000 & VIGPSO \\
& & 30 & 0.00000 & VIGPSO \\
& & 50 & 0.00000 & VIGPSO \\
& & 1000 & 0.00000 & VIGPSO \\
\cline{2-5}
& Schwefel 2.22 & 10 & 0.00000 & VIGPSO \\
& & 30 & 0.00000 & VIGPSO \\
& & 50 & 0.00000 & VIGPSO \\
& & 1000 & 0.00000 & VIGPSO \\
\hline
Part. Sep. & Dixon-Price & 10 & 0.00000 & VIGPSO \\
& & 30 & 0.00000 & VIGPSO \\
& & 50 & 0.00000 & VIGPSO \\
& & 1000 & 0.00000 & VIGPSO \\
\cline{2-5}
& Rastrigin & 10 & 0.00474 & PSO \\
& & 30 & 0.46753 & -- \\
& & 50 & 0.02656 & VIGPSO \\
& & 1000 & 0.00000 & VIGPSO \\
\hline
Non-Sep. & Rosenbrock & 10 & 0.00001 & VIGPSO \\
& & 30 & 0.00000 & VIGPSO \\
& & 50 & 0.00000 & VIGPSO \\
& & 1000 & 0.00000 & VIGPSO \\
\cline{2-5}
& Griewank & 10 & 0.69080 & -- \\
& & 30 & 0.00000 & VIGPSO \\
& & 50 & 0.00000 & VIGPSO \\
& & 1000 & 0.00000 & VIGPSO \\
\cline{2-5}
& Alpine & 10 & 0.00211 & VIGPSO \\
& & 30 & 0.09127 & -- \\
& & 50 & 0.00000 & VIGPSO \\
& & 1000 & 0.00000 & VIGPSO \\
\hline
\end{tabular}
\begin{tablenotes}
      \small
      \item Statistical significance determined using Mann-Whitney U test with $\alpha$ = 0.05. $P$-values are rounded to three decimal places.
\end{tablenotes}
\label{tab:statistical_results}

\end{table}

\begin{figure}[t!]
    \centering
    \includegraphics[width=0.45\textwidth]{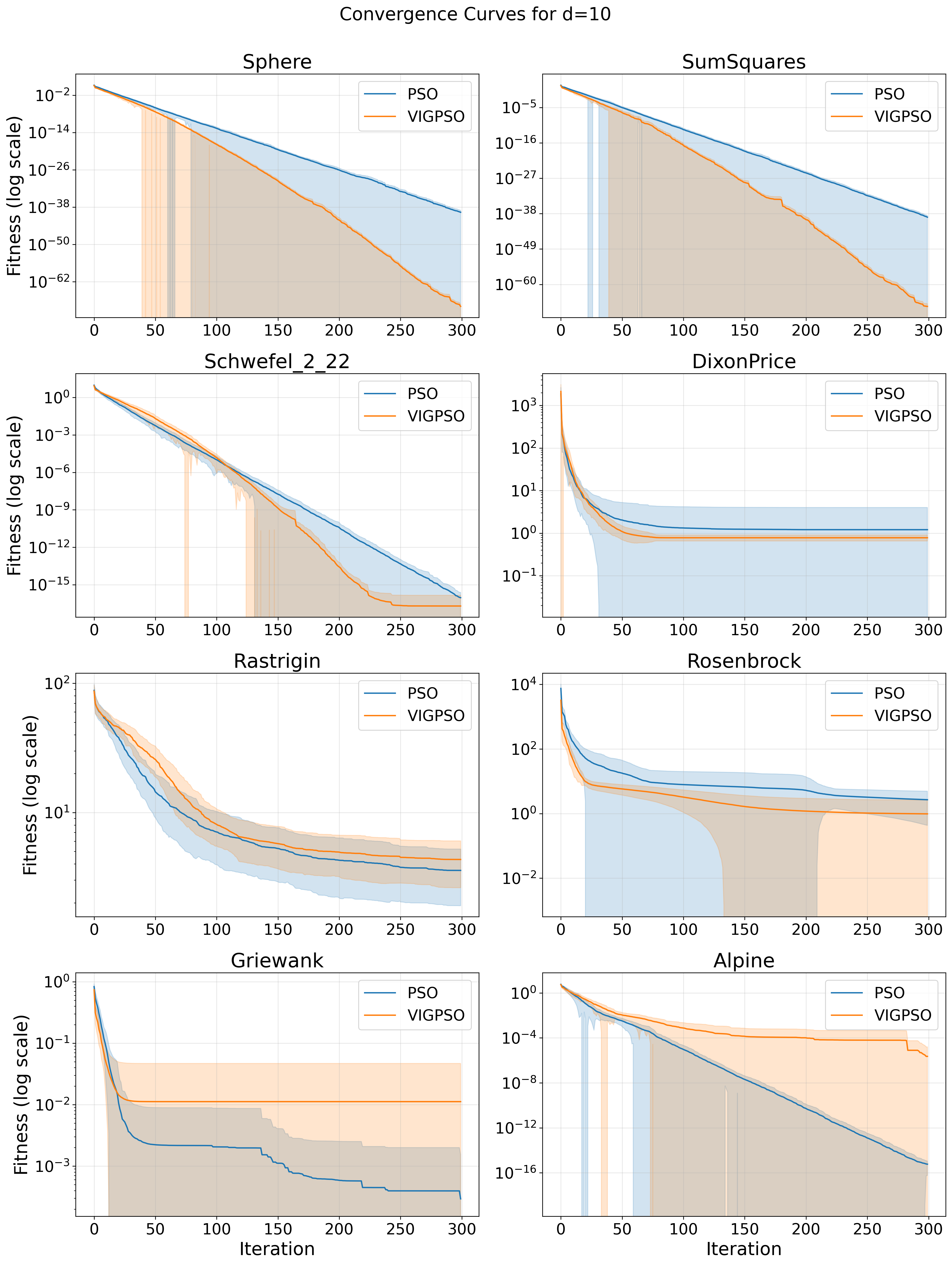}
    \caption{Mean fitness and standard deviation bands in 10 dimensions.}
    \label{fig:conv_10d}
\end{figure}

\begin{figure}[t!]
    \centering
    \includegraphics[width=0.45\textwidth]{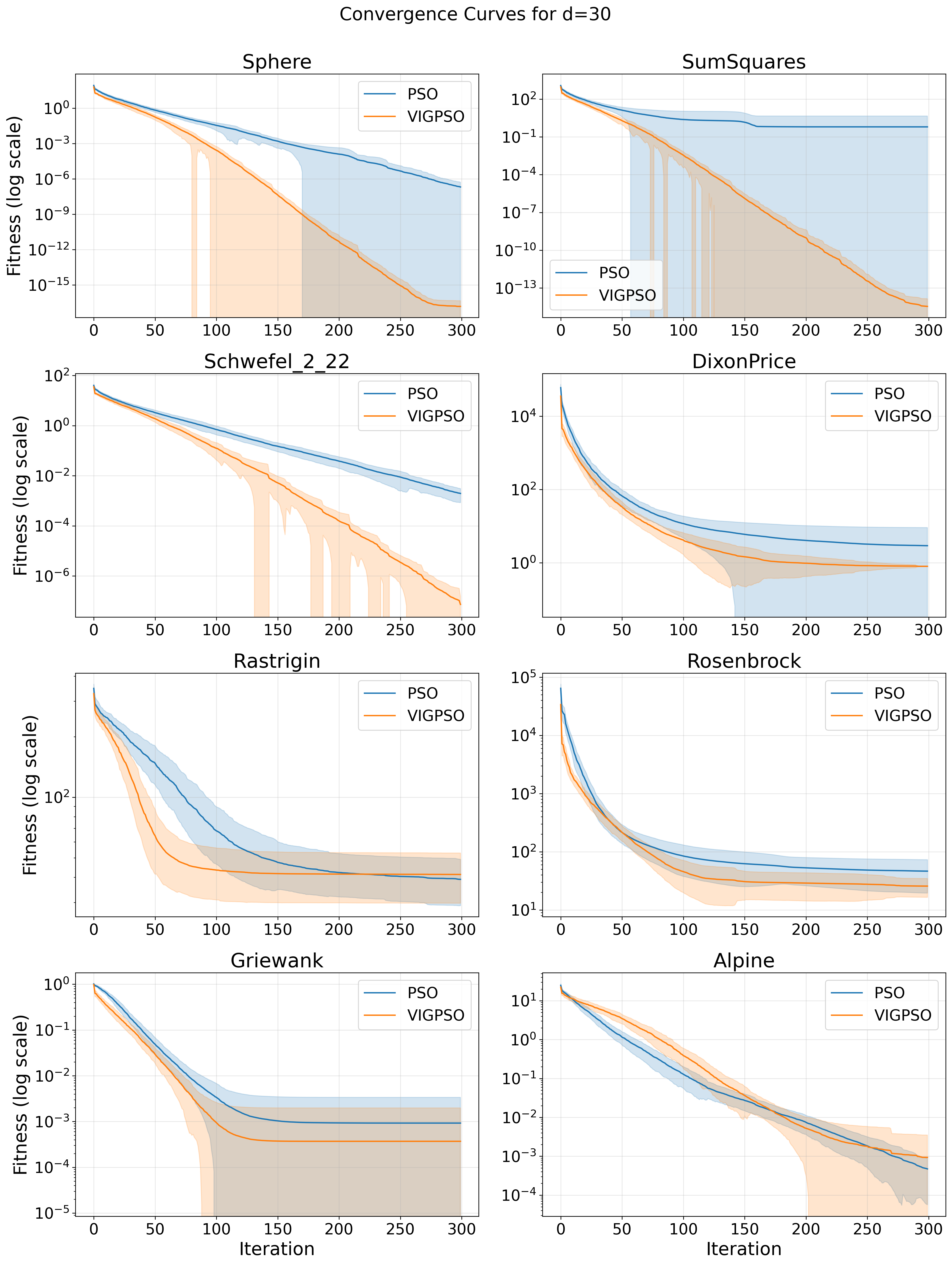}
    \caption{Mean fitness and standard deviation bands 30 dimensions.}
    \label{fig:conv_30d}
\end{figure}

\begin{figure}[t!]
    \centering
    \includegraphics[width=0.45\textwidth]{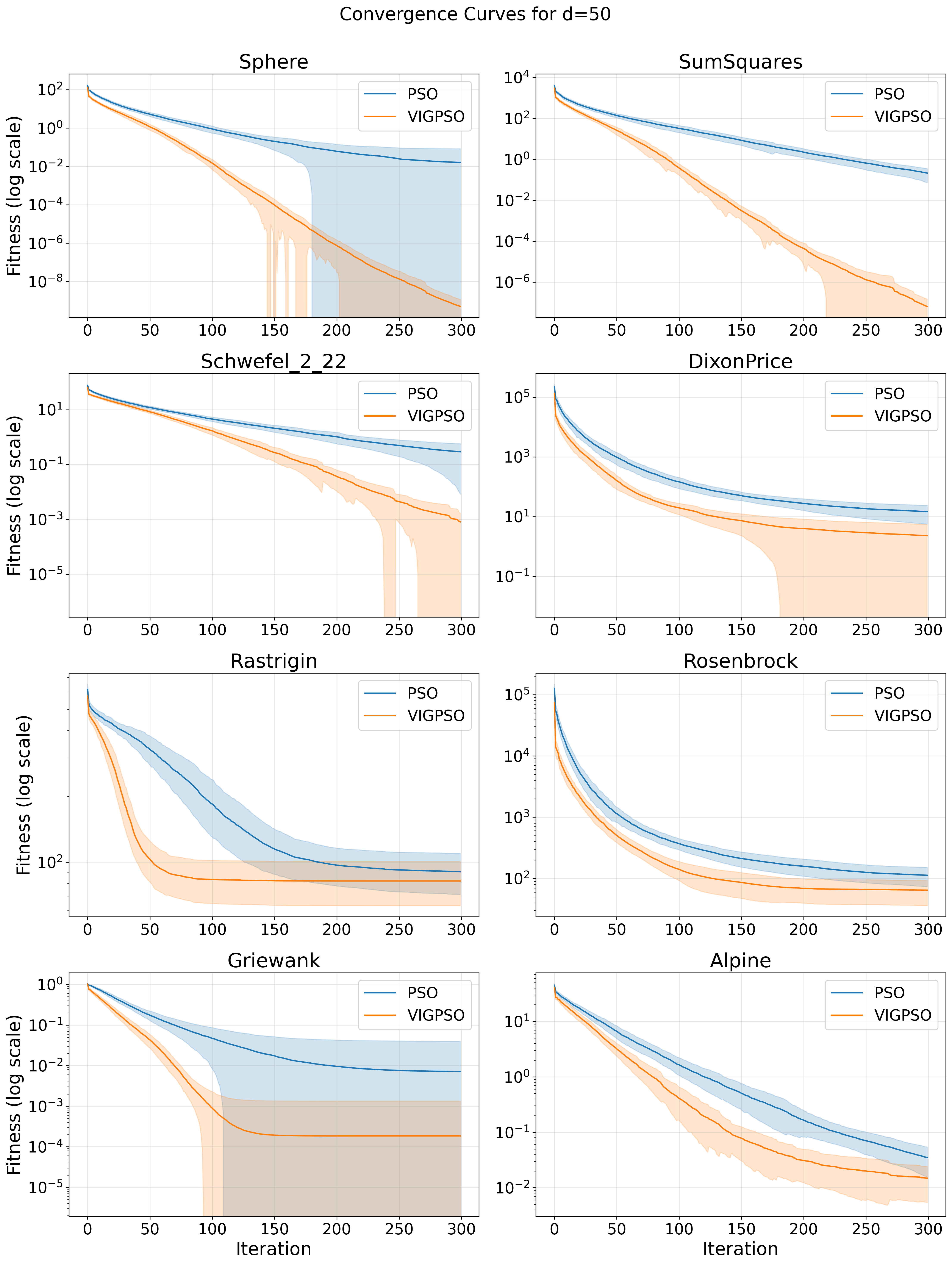}
    \caption{Mean fitness and standard deviation bands in 50 dimensions.}
    \label{fig:conv_50d}
\end{figure}

Figures \ref{fig:conv_10d}, \ref{fig:conv_30d}, and \ref{fig:conv_50d} show the convergence curves for both algorithms across all test functions at 10, 30, 50, and 1000 dimensions respectively. The solid lines represent the mean of the global best fitness values over 100 independent runs, while the shaded areas indicate one standard deviation from the mean.

\begin{figure}[t!]
  \centering
  \includegraphics[width=0.45\textwidth]{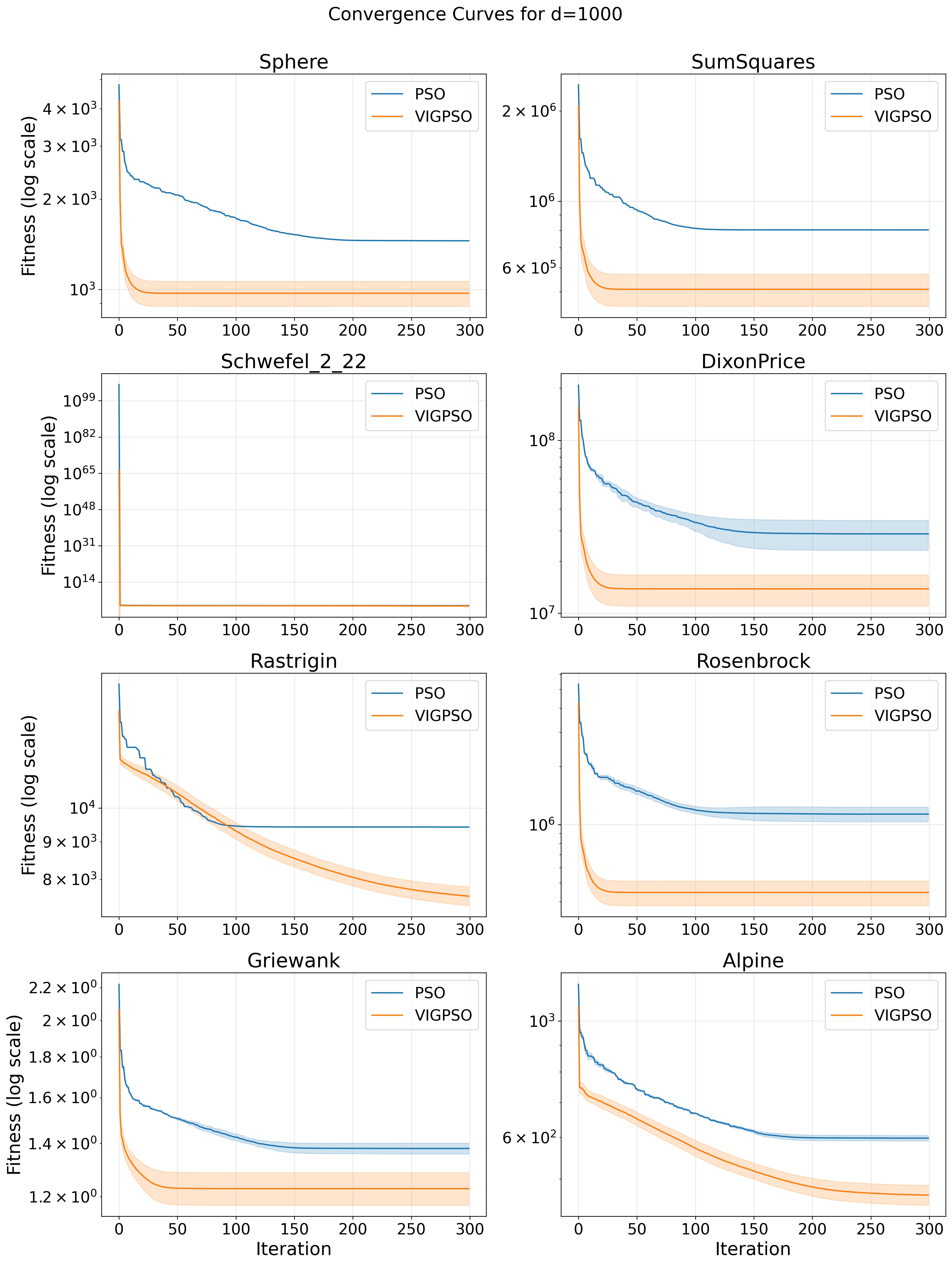}
  \caption{Mean fitness and standard deviation bands in 1000 dimensions.}
  \label{fig:conv_d1000}
\end{figure}

\begin{figure}[t!]
    \centering
    \includegraphics[width=0.45\textwidth]{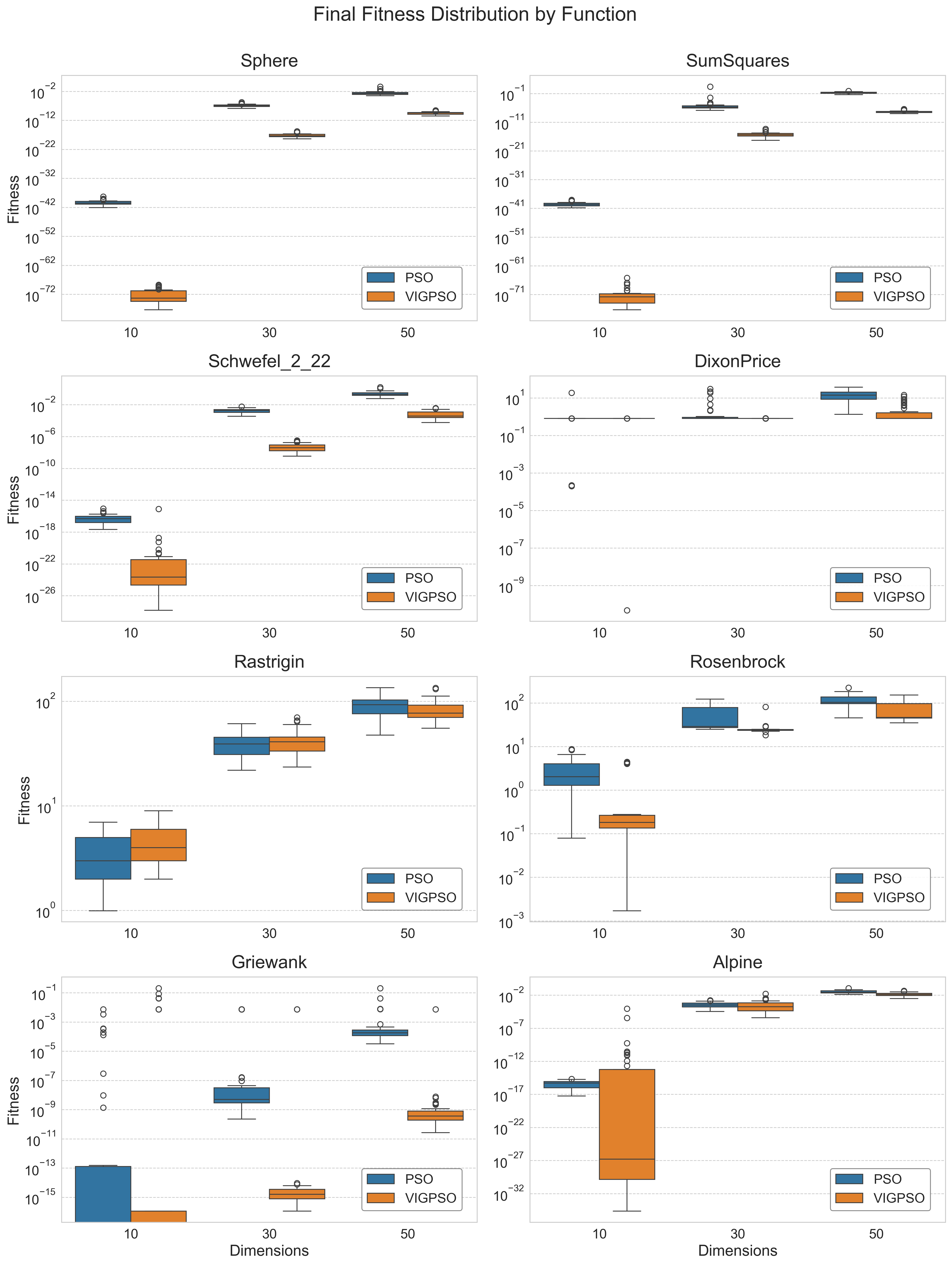}
    \caption{Distribution of final fitness values achieved by both algorithms across all test functions and lower dimensions.}
    \label{fig:boxplots}
\end{figure}

Figure \ref{fig:boxplots} shows the distribution of final fitness values achieved by both algorithms across all test functions and dimensions. The boxplots show the median, quartiles, and outliers of the final solutions obtained over 100 independent runs.

\begin{figure}[t!]
  \centering
  \includegraphics[width=0.45\textwidth]{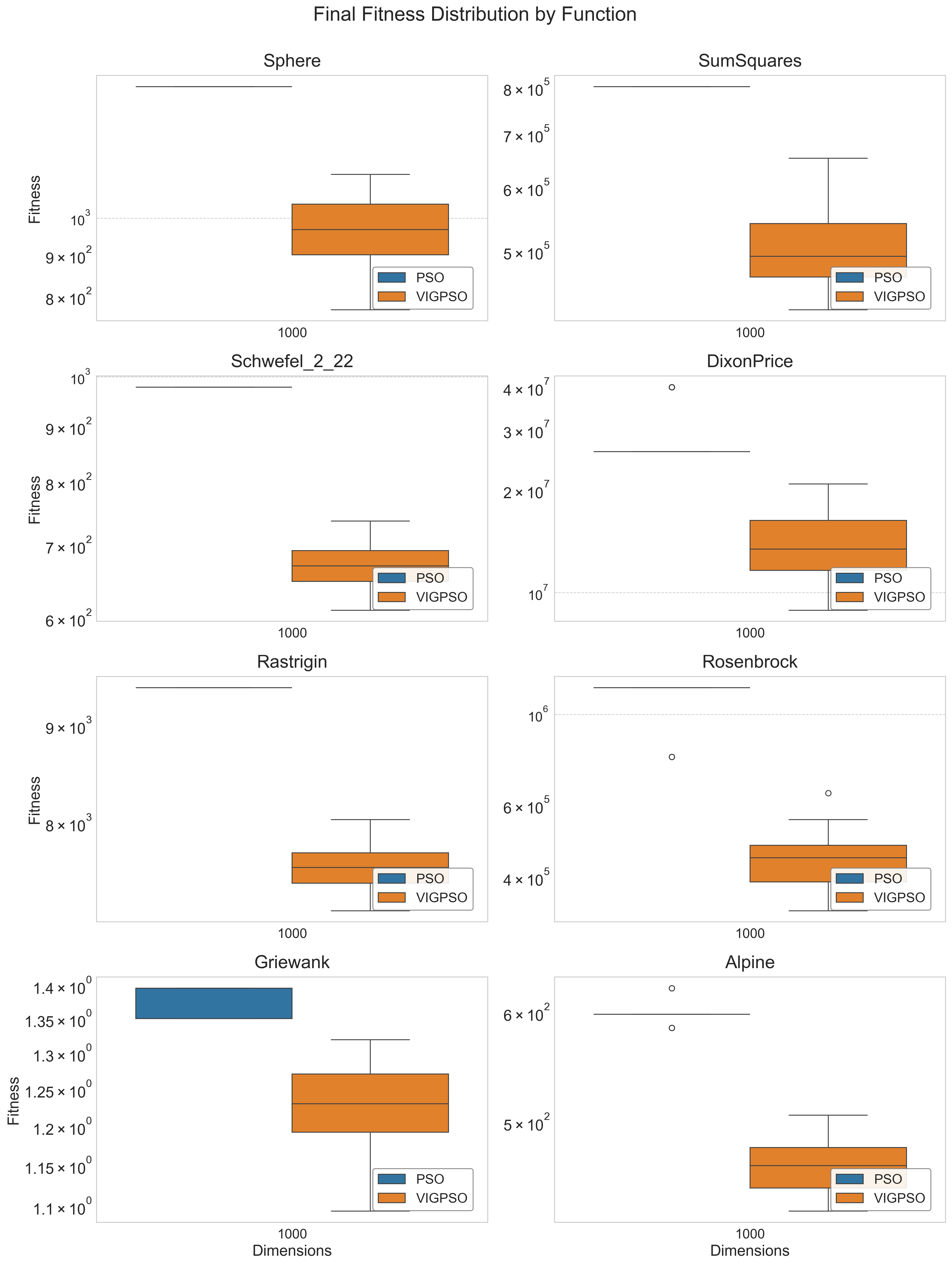}
  \caption{Distribution of final fitness values achieved by both algorithms across all test functions at 1000 dimensions.}
  \label{fig:boxplot1000}
\end{figure}

The results shown across the three performance critiera were very strong. VIGPSO was able to achieve statistically significant objective value improvements ($p<0.05$) for twenty out of the twenty-four different test configurations, as shown in Table \ref{tab:statistical_results}. This was particularly true for higher dimensions. Furthermore, three of the remaining four configurations were statistically equivalent. Standard PSO was statistically the best in only one case, and that was for a 10-dimensional problem. As dimensionality increased, the performance flipped in favor of VIGPSO. 

For fully separable functions (Sphere, Sum Squares, and Schwefel 2.22), VIGPSO showed improved convergence rates in most test cases. However, the convergence behavior was more variable for some functions, with Schwefel 2.22 showing notable fluctuations in the convergence path. It achieved faster convergence rates across most dimensionalities while maintaining smaller variance in the final solutions, as evidenced by narrower confidence bands.

For partially separable functions (Dixon-Price and Rastrigin), VIGPSO showed mixed results. Dixon-Price showed significant improvements across all dimensions ($p < 0.001$), while Rastrigin displayed more complex behavior. As mentioned above, at $d=10$, PSO performed significantly better ($p = 0.005$), at $d=30$ there was no significant difference ($p = 0.468$), and at $d=50$ VIGPSO showed superior performance ($p = 0.027$). At 1000 dimensions, VIGPSO outperformed PSO across all runs ($p < 0.001$), highlighting its scalability and adaptability as dimensionality increases. This trend underscores VIGPSO's ability to leverage variable interactions even in high-dimensional spaces.

The results for non-separable functions (Rosenbrock, Griewank, and Alpine) were also mixed. Rosenbrock showed significant improvements for VIGPSO across all dimensions ($p < 0.001$), demonstrating VIGPSO's effectiveness at navigating the function's curved valley. Griewank showed no significant difference at $d=10$ ($p = 0.691$) but significant improvements for VIGPSO at higher dimensions ($p < 0.001$ for both $d=30$ and $d=50$), with continued strong performance at 1000 dimensions. This suggests that VIGPSO's ability to capture both local and global structure becomes more valuable as the search space expands. Alpine showed improvements for VIGPSO at $d=10$ ($p = 0.002$), $d=50$ ($p < 0.001$), and notably at $d=1000$ ($p < 0.001$), reflecting the algorithm's scalability in handling functions with numerous local optima across different scales.

The results of the robustness study shown in Figure \ref{fig:boxplots} were relatively mixed when comparing VIGPSO to PSO, indicating that it has more variability. VIGPSO displayed wider interquartile ranges in approximately one third of the test configurations, in particular when the dimensionality was low. The increased variability could be attributed to the adaptive nature of VIGPSO's learning mechanism. As the algorithm builds and updates its variable interaction graph, different runs may develop different interaction patterns, leading to more diverse exploration paths.

While VIGPSO demonstrated superior optimization performance in most cases, it is important to note the additional computational overhead required for maintaining and updating the variable interaction graph. However, this overhead could be justified by the improved convergence rates and solution quality, particularly for computationally expensive optimization problems where solution quality is the primary concern.

The results support the initial hypotheses regarding improved convergence rates and solution quality. The most surprising finding was the strong performance on separable functions, where variable interactions might have been expected to be less relevant. This suggests that VIGPSO's velocity update mechanism, despite being designed for capturing variable interactions, provides beneficial optimization characteristics even when variables are truly independent. This could be due to velocity update equations introducing additional diversity in particle movements and implicit momentum effects. 

\section{Conclusion}
The results described above demonstrate VIGPSO's effectiveness in bolstering black-box optimization, as evidenced by the statistically significant improvements in 28 of 32 test configurations. A key finding was the algorithm's consistent performance across both separable and non-separable problems, with particularly strong performance as dimensionality increased.

While we achieved our primary goal of improving solution quality, the results challenge the hypothesis about optimization stability. VIGPSO showed increased variance in lower dimensions, likely due to its adaptive learning mechanism introducing additional exploration behavior. However, this trade-off between improved solutions and increased variance appears worthwhile, especially for complex problems where finding better solutions is paramount.

Future work will focus on three key areas: 1) developing adaptive mechanisms to modulate the influence of learned interactions based on problem dimensionality, 2) investigating alternative correlation metrics beyond Pearson correlation, and 3) extending the approach to constrained optimization problems. Additionally, the success of this variable interaction learning approach suggests similar mechanisms could benefit other metaheuristic algorithms. Having demonstrated that VIGs provide significant benefits in a basic PSO implementation, a valuable next step would be their incorporation into more state-of-the-art PSO variants, particularly Comprehensive Learning PSO (CLPSO) \citep{CLPSO}, which is known for its superior performance on multimodal problems.

In addition to the above, of particular interest would be adapting and evaluating VIGPSO in the context of cooperative co-evolutionary methods. For example, as detailed in Section \ref{sec:related}, methods like differential grouping are used to define subpopulations in CCEA methods but simply groups the sets of interacting variables together into disjoint factors \citep{DG-orig}. VIGPSO could be applied to each of these groups with the potential to improve performance by better utilizing the actual variable interaction structure. Furthermore, a more advanced method for CCEA has been developed, referred to as a ``factored'' evolutionary algorithm (FEA) \citep{FEA}. In FEA, the subgroups (i.e., factors) are allowed to overlap, suggesting there may be levels of interaction between groups of variables that traditional CCEA methods might miss. Furthermore, work in dynamic factors \citep{dynFEA} suggests that the factor definition could be tied directly to the adaptive construction of VIGs, as outlined here.

\bibliographystyle{IEEEtran}
\bibliography{refs}

\end{document}